\documentclass[letterpaper]{article} 
\usepackage{aaai25}  
\usepackage{times}  
\usepackage{helvet}  
\usepackage{courier}  
\usepackage[hyphens]{url}  
\usepackage{graphicx} 
\usepackage{natbib}  
\usepackage{caption} 
\usepackage{algorithm}
\usepackage{algorithmic}

\usepackage{caption, subcaption}
\usepackage{float}

\usepackage[most]{tcolorbox} 

\title{Evaluating the Evaluator: \\ Measuring LLMs' Adherence to Task Evaluation Instructions}
\author{
    Bhuvanashree Murugadoss\textsuperscript{\rm 1}, Christian Poelitz\textsuperscript{\rm 2}, Ian Drosos,\textsuperscript{\rm 2} Vu Le\textsuperscript{\rm 3}, Nick McKenna\textsuperscript{\rm 2}, Carina Suzana Negreanu\textsuperscript{\rm 2}, Chris Parnin\textsuperscript{\rm 3}, Advait Sarkar\textsuperscript{\rm 2}
}
\affiliations{
    \textsuperscript{\rm 1}Microsoft Research India
    \textsuperscript{\rm 2}Microsoft Research Cambridge UK\\
    \textsuperscript{\rm 3}Microsoft Redmond\\
    cpoelitz@microsoft.com

}

\begin{document}

\maketitle

\begin{abstract}
LLMs-as-a-judge is a recently popularized method which replaces human judgements in task evaluation~\cite{10.5555/3666122.3668142} with automatic evaluation using LLMs. Due to widespread use of RLHF (Reinforcement Learning from Human Feedback), state-of-the-art LLMs like GPT4 and Llama3 are expected to have strong alignment with human preferences when prompted for a quality judgement, such as the coherence of a text. While this seems beneficial, it is not clear whether the assessments by an LLM-as-a-judge constitute only an evaluation based on the instructions in the prompts, or reflect its preference for high-quality data similar to its fine-tune data. 
To investigate how much influence prompting the LLMs-as-a-judge has on the alignment of AI judgements to human judgements, we analyze prompts with increasing levels of instructions about the target quality of an evaluation, for several LLMs-as-a-judge. Further, we compare to a prompt-free method using model perplexity as a quality measure instead. We aggregate a taxonomy of quality criteria commonly used across state-of-the-art evaluations with LLMs and provide this as a rigorous benchmark of models as judges.
Overall, we show that the LLMs-as-a-judge benefit only little from highly detailed instructions in prompts and that perplexity can sometimes align better with human judgements than prompting, especially on textual quality.
\end{abstract}

%

\section{Introduction}
Recently, new automatic evaluation approaches that rely on LLMs have been proposed on several NLG tasks, such as summarization~\cite{liu2023geval} and machine translation~\cite{kocmi2023large}. Previous approaches~\cite{siledar2024prompt} show that for certain situations, such as when assessing textual consistency or fluency, there is high agreement between human judgements and LLM assessments, even without detailed instructions like for example how to assign specific scores. Most of these approaches prompt an LLM to give a judgement as Likert score~\cite{likert1932technique} with only simple information about the scale, e.g. ``give a judgement between 1 (bad) and 5 (good).'' More recently, LLM-based evaluations on more fine-grained task-specific criteria~\cite{ye2024flask} have also reported high agreement with human judgement, such as assessing the completeness of a solution for a question answering task. For these evaluations, often more detailed instructions are given about when to assign a specific score, similarly to rubric scoring~\cite{doi:10.3200/CTCH.53.1.27-31}.

While these results are promising for the future of automatic evaluation, it is less clear how the models achieve this agreement, and in general, it is a challenge to identify for any given task, which LLM is most appropriate to evaluate it, and with how much information, respectively how much instructions about the evaluation. Alarmingly, recent results show a clear bias in LLMs preferring their own output over others~\cite{panickssery2024llm}, and LLMs' perplexity has emerged as a possible quality criteria for filtering~\cite{ankner2024perplexed} on textual quality. This raises the question of whether some of the results on automatic evaluations with LMs reflect a model's preference for data similar to its own (high quality) fine-tuning data instead of following the provided instructions on how to measure the quality of an answer. Especially for fine-grained evaluations with detailed rubric information, we expect the instructions about when to assign a score to be adhered closely. 

In this paper we report our findings when using LLMs-as-a-judge~\cite{10.5555/3666122.3668142}, where LLMs are used as surrogates for humans judgements to evaluate several NLG and LLM-based tasks, in skill-specific settings (e.g. completeness) and skill-unspecific (e.g. textual coherence). We show the annotations for many of these quality criteria in state-of-the-art benchmarks have a high correlation with the perplexity of the LLMs, often higher than prompting the LLM for a score. We identify which evaluation settings can benefit the most from more detailed prompting and for which settings simple generic prompts, or just using models' perplexity as quality score, suffice.



In detail, we make the following three main contributions:
\begin{enumerate}
    \item We propose a novel taxonomy of qualitative evaluation criteria useful for assessing the competence of automatic evaluation methods by LLMs-as-a-judge. Our taxonomy consists of 4 evaluation categories (Content, Relevance, Integrity, and Engagement) which encapsulate 34 metrics as tested by 8 distinct state-of-the-art benchmark datasets.
    \item We systematically evaluate the effectiveness of LLMs-as-a-judge using the taxonomy with several major LLM families including GPT4, Llama3, Mistral, and Phi3 across 4 levels of increasing prompt instruction. We find that, aggregated across the taxonomy, increasing instruction by including more granular evaluation rubrics only somewhat improves the Pearson correlation of models with human judgements, by only as much as 4\%. However, some individual metrics may benefit.
    \item We evaluate the potential of simple model perplexity as an alternative automatic evaluation to LLMs-as-a-judge. While perplexity often outperforms minimal prompting in terms of correlation with human judgements when detailed rubrics are not available, textual content-related metrics are the closest aligned category. For these metrics perplexity achieves a Pearson correlation of 0.51 in contrast to 0.44 when prompting the LLM-as-a-judge, suggesting it is the better choice for simple scenarios. 
\end{enumerate}

\section{Related Work}
LLMs as evaluators for general NLG~\cite{liu2023geval}, as well as for knowledge and problem-solving tasks~\cite{ye2024flask} have been widely studied recently~\cite{chiang-lee-2023-large, li2024leveraging, gao2024llmbased}. Most previous approaches, either perform pair-wise evaluations~\cite{ji2023exploring,chen2023exploring}, measuring the preference of one of two examples for a given criterion, or perform direct assessments for a single given example and a evaluation criterion~\cite{liu2023geval, ye2024flask}. Additionally, they distinguish between reference-free evaluations, where the LLM is presented only an example and the criterion for evaluation, and reference-based evaluations with given annotated examples of different qualities or ground truth for each example.

Generally, previous works use LLMs as evaluators by using simple prompting strategies~\cite{siledar2024prompt}, only few fine-tuned models are available~\cite{kim2023prometheus, kim2024prometheus} for measuring for specific quality criteria. Most other fine-tuning approaches concentrate on scenario-specific quality feedbacks~\cite{li2023generativejudgeevaluatingalignment, wang2023shepherdcriticlanguagemodel} or on specific use-cases~\cite{mcaleese2024llmcriticshelpcatch}.

Recently, there are several approaches~\cite{liu2024llms, liu2024aligning, liu2023calibrating, stureborg2024largelanguagemodelsinconsistent} reporting biases and mismatches with human annotations, but our work is the first to study whether the models' perplexity can be a better surrogate for quality then prompting the corresponding model and whether instructions in the prompts are impacting the results across a number of different LLMs-as-a-judge.

\section{Evaluating LLMs-as-a-judge}
In this section we give a short definition of LLMs-as-a-judge and automatic evaluation using AI. We define different settings of prompting the LLMs-as-a-judge to measure the impact on the alignment of LLM judgments with human judgements. We also evaluate a prompt-free metric using simple model perplexity. This alternative approach requires no prompt engineering and transparently measures alignment with training data without bias from a prompt, so it is a compelling alternative for evaluation. Finally, we introduce a new taxonomy, aggregating the quality criteria most frequently used in state-of-the-art benchmarks for automatic evaluation with LLMs. We categorize these into 4 groups representing the major aspects of evaluating AI generated responses.

\begin{figure*}[t]
\centering 
\begin{tcbraster}[raster columns=3, raster equal height, raster column skip=0.1cm]  
  \begin{tcolorbox}[title=2. Generic prompt, colback=white, sharp corners]  
\tiny
\# Task to evaluate\newline
Your tasks is to evaluate the interaction between a user and an AI assistant. I want you to evaluate
the assistant's response. Evaluate the quality of the response. \newline
The AI responses are for: \{task description\}.\newline
\newline
\# Sample to evaluate\newline
\{example\}\newline
\newline
\# Instructions\newline
Evaluate the quality of the response from the sample and return a score between 1 (bad) and 5 (very good) as:\newline
\#\# Score: [Number] 
  \end{tcolorbox}  
  \begin{tcolorbox}[title=3. Criteria specific prompt, colback=white, sharp corners]  
\tiny
\# Task to evaluate\newline
Your tasks is to evaluate the interaction between a user and an AI assistant. I want you to evaluate
the assistant's response. Evaluate the response for the given rubric below. Use the rubric to guide you evaluating and base all your evaluation decision on the rubric. \newline
The AI responses are for: \{task description\}.\newline
\newline
\# Sample to evaluate\newline
\{example\}\newline
\newline
\# Rubrics \newline
Logicality\newline
\newline
\# Instructions\newline
Evaluate the quality of the response from the sample and return a score between 1 (bad) and 5 (very good) as:\newline
\#\# Score: [Number]  
  \end{tcolorbox}  
  \begin{tcolorbox}[title=4. Full rubric prompt, colback=white, sharp corners]  
\tiny
\# Task to evaluate\newline
Your tasks is to evaluate the interaction between a user and an AI assistant. I want you to evaluate
the assistant's response. Evaluate the response for the given rubric below. Use the rubric to guide you evaluating and base all your evaluation decision on the rubric. \newline
The AI responses are for: \{task description\}.\newline
\newline
\# Sample to evaluate\newline
\{example\}\newline
\newline
\# Rubrics \newline
Logicality: Measure how much the story obeys your commonsense.\newline
Score 1: The story is full of absurd things. \newline
Score 2: The story has one or two things make sense, but generally very absurd.\newline
Score 3: The story roughly makes sense.\newline
Score 4: The story largely makes sense, except one or two things.\newline
Score 5: The story totally complies with commonsense. \newline
\newline
\# Instructions\newline
Evaluate the quality of the response from the sample and return a score between 1 and 5 as:\newline
\#\# Score: [Number]  
  \end{tcolorbox}  
\end{tcbraster}  
\caption{Our prompting settings. We measure how much influence the information about the actual evaluation has for model performance as LLM-as-a-judge. For setting 1, perplexity, we don't prompt the models but calculate the models' perplexity for the task solution in the example instead. The example prompts shown above are used for the LLMs-as-a-judge to measure the quality for the criterion \textbf{logicality} as defined in for the benchmark dataset \textbf{TheNextChapter}.} 
\label{fig:prompt_examples}
\end{figure*}  

\subsection{LLMs-as-a-judge}
As LLM-as-a-judge we refer to the definition introduced by \citep{10.5555/3666122.3668142} as potential replacement for human annotations by prompting an LLM for a judgement of an AI assistant response. We concentrate on judging textual examples only e.g., AI generated summaries for news articles~\cite{fabbri2021summeval} or step-by-step solution to mathematical reasoning questions~\cite{golovneva2023roscoe}. We phrase the task to judge an AI generated response as the following: 
Given a task A and an AI generated solution B, judge the quality of the solution B considering only the task A. In contrast to other previous approaches, we perform a reference-free evaluation where we do not provide a possible correct reference solution. We solely rely on the models' ability to judge the solution given only the task.

To measure the impact of prompting the LLM-as-a-judge, we study the LLMs' performance in \textbf{4} different settings:
\begin{enumerate}
    \item \textbf{Perplexity}: We score each task solution by its perplexity under the corresponding LLM, given only the task description. This approach is unbiased by prompts, so it transparently measures alignment with model training data, providing a good comparison and alternative to prompt-based approaches.
    \item \textbf{Generic quality prompt}: We prompt each LLM-as-a-judge with a basic instruction to measure the quality of the task solution, but give no specific criteria or instructions. In this case, we rely solely on the models' prior knowledge about the quality for the task solution from the examples generated.
    \item \textbf{Criteria specific prompt}: We prompt each LLM-as-a-judge with an instruction to measure the quality for a specific criteria e.g., \textbf{coherence}. We only provide the name of the criteria, not a definition. We rely on the models' prior knowledge of the specific quality criteria only.  
    \item \textbf{Full rubric prompt}: We prompt each LLM-as-a-judge with an instruction to measure the quality for a specific quality criterion, together with a definition of the criterion and instructions when to assign each rubric score e.g., ``\textit{Score 1: Incoherent text with many logical flaws}.''
\end{enumerate}

We evaluate different LLMs-as-a-judge under the above settings on several different benchmarks (as described in the next subsection). We use the criteria as specified in the corresponding annotation guidelines from the benchmark datasets. For setting 4, we use all available annotation guidelines with information about the criteria and when to assign each score. We extract this information directly from benchmarks into a full rubric containing information about the criteria and the scores. For our experiments, we structure the settings from least instructive (Perplexity / no prompting) to most instructive (Full rubric information with instructions when to assign a score). In Fig.~\ref{fig:prompt_examples} we show the different setting of prompting for an example quality criterion.

\subsection{Datasets}
We use 8 different open-source benchmark datasets commonly used for LLM-based evaluations with human annotations for several evaluation criteria per task. The datasets contain task which span several aspects from coarse-grained NLG-quality evaluations, to fine-grained very task specific evaluations with detailed information about how to score examples.

Firstly, we leverage two of the most prominently used datasets for coarse-grained NLG-quality evaluations: The \textbf{SummEval}~\cite{fabbri2021summeval} dataset contains news article summaries generated by different models together with human annotations for 4 different quality criteria e.g., fluency; and the \textbf{TopicalChat}~\cite{gopalakrishnan2019topical} dataset contains human conversations over 8 different topics annotated by humans for 5 different quality criteria e.g., engagement. Further, we use two more challenging benchmark datasets for coarse-grained NLG-evaluations: the \textbf{OpinSummEval}~\cite{shen2023opinsummevalrevisitingautomatedevaluation} dataset is a opinion summarization dataset, which consists of review summaries annotated for aspects, opinions and sentiments; the \textbf{InstruSumm}~\cite{liu2023benchmarkinggenerationevaluationcapabilities} dataset, consists of news article summaries following specific instructions with human annotations for content specific quality-criteria e.g., amount of missing information. 

Second, we use two benchmark datasets for more fine-grained NLG evaluations: the \textbf{Hanna}~\cite{chhun-etal-2022-human} dataset and the \textbf{TheNextChapter}~\cite{xie2023chapterstudylargelanguage} dataset contain creative stories generated for a given initial user prompt. Each story is annotated by humans for NLG and style based criteria e.g., coherence, but also for more unconventional criteria like surprise.
Finally, we use two task-specific evaluation benchmark datasets with quality-criteria depending in task solution quality: \textbf{Roscoe}~\cite{golovneva2023roscoe} is a collection of datasets of reasoning tasks, together with GPT3 generated with step-by-step solutions. The human annotations cover coarse-grained task specific evaluation criteria like ``missing step''; the \textbf{Flask}~\cite{ye2024flask} dataset contains several knowledge and problem solving tasks with LLM generated solutions. The human annotations cover more fine-grained task-specific criteria like completeness and factuality. Most criteria need an understanding of the solution e.g., completeness.

\subsection{Criteria taxonomy}
\begin{figure}[t]
    \centering\includegraphics[width=0.95\linewidth, height=0.95\linewidth]{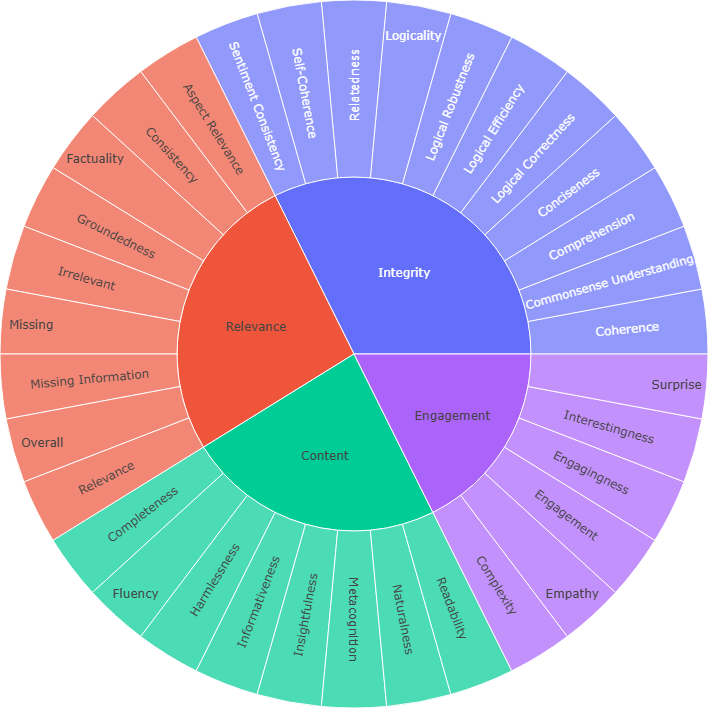}
\caption{Taxonomy of quality criteria summarizing current state-of-the-art benchmark datasets and criteria used for automatic evaluations with LLMs. We group all 34 quality criteria as defined in the 8 different benchmark datasets into 4 groups: \textbf{Content-based, Engagement-based, Integrity-based, Relevance-based criteria}.}
\label{taxonomy}
\end{figure}
We introduce a simple taxonomy of quality evaluation criteria based on current state-of-the-art benchmark datasets and quality criteria commonly used for automatic evaluations by LLMs. We define 4 groups of quality criteria, relevant for automatic evaluation:
\begin{enumerate}
    \item \textbf{Content-based criteria}: Measure how well the solution is presented to the user, for example, whether a news article summary is fluent.
    \item \textbf{Engagement-based criteria}: Measure how engaging the solution is, for example, whether a generated story contains an element of surprise.
    \item \textbf{Integrity-based criteria}: Measure how consistent and logical coherent the solution is, for example, whether a math solution is correct.
    \item \textbf{Relevance-based criteria}: Measure how relevant the solution is for the given task, for example, whether a legal advice answer contains irrelevant information.
\end{enumerate} 

We assign each of the criteria used in the benchmark datasets above to these 4 groups (Fig.~\ref{taxonomy}). For \textbf{content-based criteria}, we are interested in how to measure the quality of the content as it is presented to the user. This includes mainly criteria of the textual quality of the solution. For example, the criterion \textbf{fluency} is used for measuring the quality of the summaries in the SummEval dataset and hence is a content-based criteria. The \textbf{engagement-based criteria}, combine criteria of how the AI generated solution engages with the user. This includes for example the \textbf{empathy} criterion used to measure the quality of the generated stories in the Hanna dataset. The remaining two groups concentrate on more task specific evaluation criteria. \textbf{Integrity-based criteria} measure the coherence of task solution and whether it makes sense logically. For example the criterion \textbf{logical correctness}, used for measuring the quality of (mathematical) resonsning or coding task solutions in the Flask dataset, is a integrity-based criterion. Finally, \textbf{relevance-based criteria} measure the direct relevance of a task solution to the actual task. This includes for example the criterion \textbf{relevance}, used for measuring the connections of task solutions and initial task in several of the benchmark datasets e.g., TheNextChapter dataset. The separation of the quality criteria classes are not a 100\% perfect and there are overlaps, for example content-based criteria like \textbf{readability}, can also be seen as engagement-based, since less legible solution are also less engaging. 

\subsection{Model Selection for LLM-as-a-judge}
To understand how model size and finetuning affect performance across the different quality criteria and settings of prompting, we test several current LLMs: GPT4-Turbo~\cite{openai2024gpt4}-0125 as large closed-model baseline; Llama3 70b~\cite{touvron2023llama, dubey2024llama3herdmodels} as a medium size open-model; Llama3 8b, Mistral-v0.3~\cite{jiang2023mistral} as small open-models,  Phi3~\cite{abdin2024phi3}-Medium-128k as fine-tuned model for reasoning, and Prometheus-2~\cite{kim2024prometheus} as fine-tuned models for evaluation tasks\footnote{In all experiments, we generate 20 responses with a temperature of 0.3 and use the average of all generated scores, similar to~\cite{liu2023geval}.}. 

\section{Results}
In this section, we present the main results of the evaluations using the different LLMs-as-a-judge under the different settings of prompting. Analogous to previous work \cite{liu2023geval}, to measure the quality of the evaluations we calculate the Pearson correlation of the generated scores by the LLMs-as-a-judge, respectively the perplexity values, and the human annotations given for each quality criteria from the benchmark dataset. 
We split the results section into model level, dataset level and criteria level results. In the model level results subsection, we present the results comparing the different LLMs under the setting of prompting, averaging over all criteria; the dataset level results subsection presents the results when we compare the different datasets under the setting of prompting average over all criteria; the criteria level results subsection presents the results when we compare models and setting of prompting under the different groups of criteria. Finally, we present the results of a detailed analysis for each group of criteria from the introduced taxonomy. 

\subsection{Model level results}

\begin{table}[t]
\centering  
\small
\begin{tabular}{l|l|ll|l}  
Setting & 1 & 2 & 3 & 4 \\
\hline
GPT4-Turbo & - & 0.414 & 0.468 & \textbf{0.469} \\
GPT3.5-Turbo & - & 0.269 & 0.310 & \textbf{0.313} \\
Llama3 70b & 0.295 & 0.299 & 0.349 & \textbf{0.367} \\
Phi3-Medium  & 0.289 & 0.324 & \textbf{0.367} & 0.334 \\
Llama3 8b & 0.288 & 0.256 & 0.294 & \textbf{0.352} \\
Mistral & \textbf{0.324} & 0.261 & 0.259 & 0.311 \\
Prometheus-2 & \textbf{0.333} & - & - & 0.266 \\
\end{tabular}  
\caption{Pearson correlations of the scores generated by different LLMs-as-a-judge with the human annotations from the different datasets for the different settings (\textbf{1 - Perplexity / No Prompting, 2 - Generic prompt, 3 - Specific prompt, 4 - Full rubric}). Bold numbers show highest agreement with human annotations under the setting for each model.}
\label{tab:models_metrics_correlations}
\end{table}

\textbf{There is only small effect adding full rubric information.} Providing the LLMs-as-a-judge with more detailed rubric information of the quality criteria, generally has only small influence on evaluation performance for the large and mid-size models, and might even be disadvantageous in certain situations (see Tab.\ref{tab:models_metrics_correlations}). For instance, Phi3's performance decreases when complete rubric details are provided compared to simple prompts which only mention the criterion name in the prompt. Here, Phi3's prior knowledge about evaluating the criteria has higher agreement with human annotators compared to when using full rubric information. Only the smaller Llama3 8b and Mistral models see improvements when given comprehensive rubric information for assessment. Among the open models, Llama3, both the 70b and 8b versions, perform best. Meanwhile, Mistral and Prometheus-2 do not show improvements when the LLM is prompted, with models' perplexity having higher correlation then the generated scores. For Prometheus-2, we only report perplexity and full rubric information in the prompts since this aligns with the fine-tuning data for this model and both setting 2 and 3 did return very poor results. 

\textbf{GPT4 performs best among all models.} As may be expected, prompting GPT4-as-a-judge, even for a generic quality judgement, results in the highest performance in terms of agreement with human annotations compared to all other models tested. Further, GPT4's judgements do only improve marginally from prompting setting 3 to 4, indicating that GPT4's prior knowledge about evaluating does already agree with the human judgements to a high degree without the need to add more detailed rubric information about the evaluation.


\subsection{Dataset level results}

\begin{table}[t]
\centering  
\small
\begin{tabular}{l|l|ll|l}  
Setting & 1 & 2 & 3 & 4 \\
\hline
Flask & \textbf{0.448} & 0.365 & 0.409 & 0.408 \\
Hanna & 0.237 & 0.232 & 0.262 & \textbf{0.318} \\
TheNextChapter & 0.275 & 0.193 & 0.273 & \textbf{0.340} \\
Summeval & \textbf{0.408} & 0.345 & 0.369 & 0.376 \\
TopicalChat & 0.189 & 0.421 & \textbf{0.426} & \textbf{0.426} \\
InstruSum & 0.160 & 0.130 & \textbf{0.163} & 0.153 \\
OpinSummEval & 0.179 & 0.316 & \textbf{0.342} & 0.328 \\
Roscoe & 0.159 & 0.294 & 0.349 & \textbf{0.392} \\
\end{tabular}  
\caption{Pearson correlations of the scores generated by different LLMs-as-a-judge with human annotations from the \textbf{InstruSum} dataset for each setting (\textbf{1 - Perplexity / No Prompting, 2 - Generic prompt, 3 - Specific prompt, 4 - Full rubric}). Bold numbers show highest agreement with human annotations in the setting for each dataset.}
\label{tab:datasets_metrics_correlations}
\end{table}

\textbf{Perplexity correlates with text quality criteria.} We observe (see Tab.~\ref{tab:datasets_metrics_correlations}) that the quality criteria from datasets with simple textual content creation tasks e.g., summarization in the SummEval dataset or story generation in the Hanna dataset, show high agreement with models' perplexity compared to simple prompting (setting 2 and 3). For more complex NLG tasks, which depend on several aspects and multiple possible steps, the human annotation correlate less strongly with perplexity compared prompting the LLMs-as-a-judge with more information. For example the opinion summary evaluations from the OpinSummEval dataset uses criteria which depend on sentiment identification and extractions of the key aspects, in these cases prompting the LLM seems necessary. 

\textbf{Full rubric information helps for non-default textual quality evaluations}. Unusual textual quality evaluation tasks which measure the quality beyond simple textual criteria like fluency, can benefit from more full rubric information about the the evaluation task. For example, we observe that for the TheNextChapter dataset, full rubric information in the prompts to the LLMs-as-a-judge leads to judgements with the highest correlations with human annotations. Compared to other datasets for textual quality evaluation, these two datasets contain much more complex texts e.g., creative stories with non-default quality criteria like relatedness which is difficult to estimate without additional information by an LLM.
Furthermore, evaluating more complex tasks which include more than text quality, like the logical reasoning tasks, benefit also from more detailed rubric information in the prompts. The logical and mathematical reasoning tasks in the Roscoe datasets for example do benefit from more information to effectively judge as shown by the higher correlations with the human judgements compared to prompting with less information or using perplexity.

\textbf{Dataset level analysis can be misleading.} Models' perplexity on both the Flask dataset and the SummEval dataset, outperforms simple prompting in aligning to human judgements. While the quality criteria in the SummEval dataset primarily focuses on textual quality where we expect perplexity to perform well for example, the Flask dataset consist a variety of different quality criteria which make it difficult to generalise and the average correlations values might be biased the high values on the text related criteria. In the next subsection, we investigate this issue by using the previously introduced taxonomy to analyze results on a per-criteria class basis rather than average results per dataset. 

\subsection{Criteria level analysis}

\begin{figure}[t]
    \centering
        \includegraphics[width=0.9\linewidth, height=0.9\linewidth]{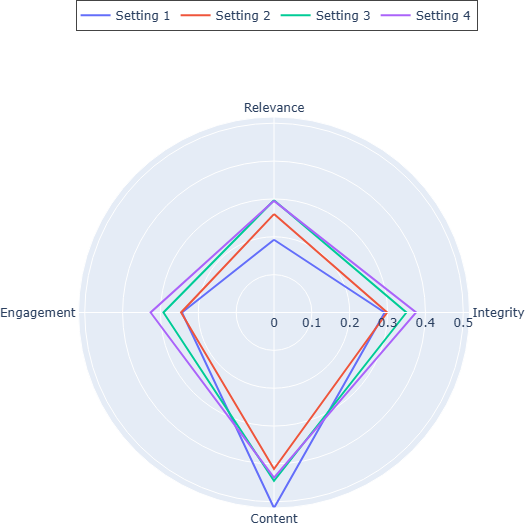}
\caption{Radar chart of average Pearson correlations for the quality criteria groups for each of different settings of prompting (\textbf{1 - Perplexity / No Prompting, 2 - Generic prompt, 3 - Specific prompt,  4 - Full rubric}) over all models.}
\label{fig:taxonomy_settings_corr}
\end{figure}

\textbf{Content-based quality criteria correlate the most with perplexity.} When evaluating quality with a focus on textual context, perplexity seems a viable alternative to prompting LLMs-as-a-judge. We observe that on average the agreement with human annotations is more then 20\% higher when using models' perplexity to judge the quality compared to prompting (Fig.~\ref{fig:taxonomy_settings_corr}). Further, there are only small differences between using a simple generic quality prompt for evaluation compared to all other settings of prompting, showing that models' prior knowledge generates judgements with high agreement with human judgements on textual quality. 

\textbf{Engagement-based quality criteria benefit the most from full rubric information.} These criteria are unconventional as they assess the likelihood of a user feeling personally engaged, as opposed to merely evaluating straightforward text quality. Access to full rubric information can help judging with directives, particularly when the evaluation is more unusual and different from the text quality alone.

\textbf{Often, there is only little improvement of adding full rubric information for most criteria groups.} Except for the engagement-based criteria, there is only limited effect on adding full rubrics information to the prompts. Further, simple prompts for generic quality judgements results in similar correlation values with the human annotations than detailed information for content and relevance based evaluation criteria. This confirms again that advanced models' prior knowledge of text quality or relevance already has a high agreement with human judgements.

\begin{figure}[t]
    \centering
        \includegraphics[width=0.9\linewidth, height=0.9\linewidth]{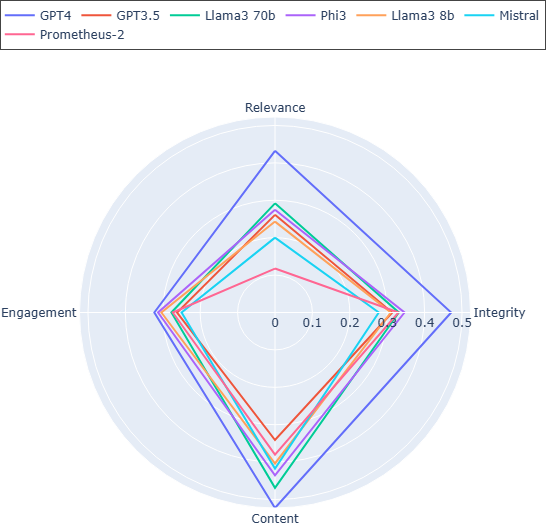}
\caption{Radar chart of average Pearson correlations for the quality criteria groups for each of the different LLMs-as-a-judge over all setting of prompting.}
\label{fig:taxonomy_models_corr}
\end{figure}

\textbf{GPT4 clearly outperforms all other models.} GPT-4 particularly outperforms in relevance and integrity quality criteria, surpassing other models on these criteria (Fig.~\ref{fig:taxonomy_models_corr}). Although Phi3 matches GPT-4's performance in engagement based criteria, it generally performs less consistent with human annotations across other evaluative measures; Mitral's performance falls short in criteria associated with relevance and integrity; LLamA3-70b exhibits a marginally improved performance over Phi3 when it comes to content and relevance-based criteria.

\subsection{Details on content-based criteria results}


\begin{table}[t]
\centering  
\small
\begin{tabular}{l|l|ll|l}  
Setting & 1 & 2 & 3 & 4 \\
\hline
harmlessness & \textbf{1.000} & 0.486 & 0.749 & 0.928 \\
completeness & \textbf{0.764} & 0.708 & 0.707 & 0.707 \\
readability & \textbf{0.731} & 0.256 & 0.329 & 0.341 \\
fluency & \textbf{0.430} & 0.310 & 0.314 & 0.312 \\
metacognition & 0.374 & \textbf{0.454} & 0.436 & 0.330 \\
insightfulness & 0.266 & 0.327 & \textbf{0.444} & 0.339 \\
naturalness & 0.274 & 0.454 & 0.458 & \textbf{0.466} \\
\end{tabular}  
\caption{Pearson correlations of scores generated by different LLMs-as-a-judge with human annotations for content-based evaluation criteria, from the benchmark dataset for respective settings (\textbf{1 - Perplexity (No Prompt), 2 - Generic prompt, 3 - Specific prompt,  4 - Full rubric})}
\label{tab:content_based_metrics_correlations}
\end{table}

Drilling down the content-based evaluations criteria, we observe that perplexity outperforms prompting mainly on structural text quality criteria. For example, human annotations for \textbf{fluency} from the SummEval dataset have much higher agreement with perplexity compared to all prompting methods. This quality criterion judges grammar, spelling, and sentence structure for example. 
On the other hand, evaluating for more complex, specific and subjective content-based criteria like \textbf{naturalness}, which measures how natural the task response sounds, benefits from more instructions in the prompts for the LLMs-as-a-judge. Here, instructions to judge how much the task solution resembles a human answer improves agreement with human judgements.

Notably, we observe that model perplexity has 100\% agreement with the human annotations for \textbf{harmlessness} on Flask datasets. This might reflect the strong influence fine-tuning has in inhibiting the generation of harmful content~\cite{dubey2024llama3herdmodels}. Conversely, prompting for measuring harmfulness performs much lower until we provide full rubric information in the prompts. 

\subsection{Details on engagement-based criteria results}
\begin{figure}[t]
    \centering
        \includegraphics[width=0.9\linewidth]{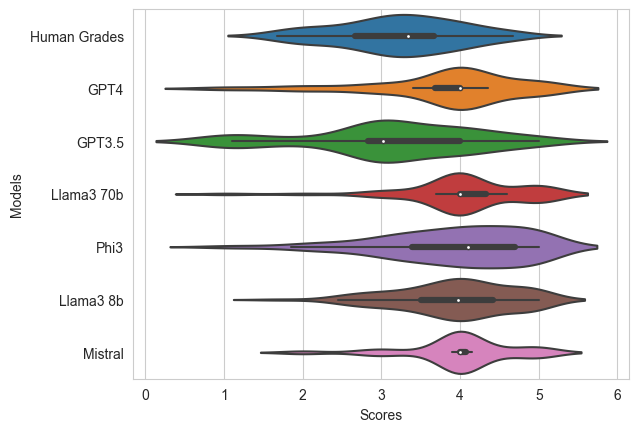}
\caption{Violin plot of the generated scores by the LLMs-as-a-judge for the engagement-based criterion \textbf{empathy}, together with the corresponding human annotations.}
\label{fig:score_distributions_violin_hanna_empathy}
\end{figure}

Engagement-based criteria are more challenging to judge since they are often subjective. We observe that there are fewer performance differences between the models compared to the results on the other criteria e.g., GPT4 judgements have an agreement (by Pearson correlation) of 0.32, Phi3 of 0.31 and Llama3 8b of 0.3.
The overall performance of all models on these criteria is lower and the scores generated by the models have higher variance compared to other criteria. Further, human annotations for these criteria have on average lower scores with higher variance. For example, for the criterion \textbf{empathy} from the Hanna dataset (Fig.~\ref{fig:score_distributions_violin_hanna_empathy}), human annotations show a significant degree of variation, and Phi3 notably generates scores with higher variability, unlike most other models that tend to cluster around a singular score. Hence, the distribution in engagement-based criteria may explain why Phi3 outperforms other models.

\subsection{Details on relevance-based criteria results}

\begin{table}[t]
\centering  
\small
\begin{tabular}{l|l|ll|l}  
Setting & 1 & 2 & 3 & 4 \\
\hline
GPT4-Turbo & - & 0.363 & \textbf{0.365} & 0.318 \\
GPT3.5-Turbo & - & 0.227 & 0.242 & \textbf{0.274} \\
Llama3 70b & 0.084 & \textbf{0.301} & 0.288 & \textbf{0.301} \\
Phi3 & 0.011 & 0.260 & \textbf{0.263} & 0.249 \\
Llama3 8b & 0.007 & 0.180 & \textbf{0.226} & 0.180 \\
Mistral & 0.108 & \textbf{0.154} & 0.126 & 0.102 \\
\end{tabular}
\caption{Pearson correlations of the scores generated by different LLMs-as-a-judge with the human annotations for \textbf{groundedness} for the different setting (\textbf{1 - Perplexity / No Prompting, 2 - Generic prompt, 3 - Specific prompt, 4 - Full rubric})} 
\label{tab:groundedness_settings_correlations}
\end{table}

For relevance-based quality criteria, we assume that the models need robust information about the problem to measure whether the information in the task solution is relevant, and estimate to what degree. Models' perplexity, but also generic prompts seem not sufficient for evaluation since relevance is more specific to certain aspects of the task solution. Still, we also observe that including a full rubric doesn't always appear necessary; instead the size of the models seem more important. For the criterion \textbf{groundedness} from the TopicalChat dataset for example (Tab.~\ref{tab:groundedness_settings_correlations}), we identify a clear trend of increasing agreement with the human judgements with larger models as LLM-as-a-judge.

\subsection{Details on integrity-based criteria results}
\begin{figure}[t]
    \centering
        \includegraphics[width=0.9\linewidth]{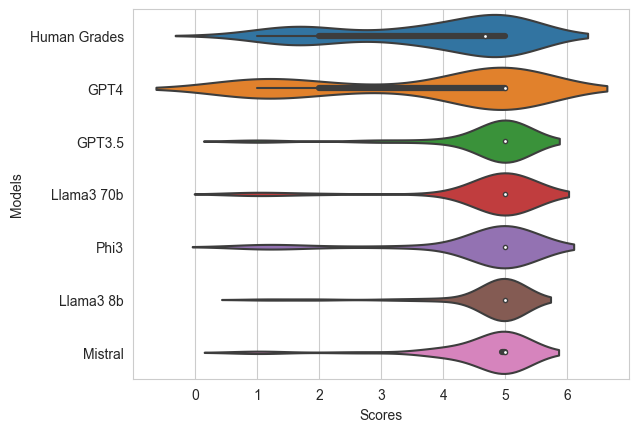}
\caption{Violin plot of the generated scores by the LLMs-as-a-judge for integrity-based criterion \textbf{logical correctness} from the Flask dataset, together with human annotations.}
\label{fig:score_distributions_violin_flask_logical_reasoning}
\end{figure}

Similar to the relevance-based quality criteria, we assume that to measure the quality for integrity-based criteria the LLMs-as-a-judge need to have an understanding of the task, but also the ability to solve the task itself. We hypothesize that, for task-specific evaluations, the underlying LLM-as-a-judge actually needs to be able to solve the task itself to apply the correct score. As reported in ~\cite{lin2024criticbenchbenchmarkingllmscritiquecorrect}, LLMs' ability to critic a task solution correlates with its ability to solve the task. 

We exemplify this by the evaluations for the integrity-based evaluation criterion on the criterion \textbf{logical correctness} from the Flask dataset. This criterion reflects the correctness elements which are reflected in human annotations, which are more clearly scored as high (for logically correct) or low (for logically incorrect) scores. To select the appropriate scores, the LLM-as-a-judges need to know what is correct and what is wrong. Here, GPT-4 significantly outperforms all other models by a wide margin with a Pearson correlation of 0.68 with the human judgements in contrast to 0.34 for Llama3 70b and 0.33 for Phi3, for example.

To illustrate this, we plot the generated scores of the LLMs-as-a-judge (Fig.~\ref{fig:score_distributions_violin_flask_logical_reasoning}) and the human annotations. We observe that only GPT4 is able to generate lower scores to judge a task solution as ``bad.'' All other models predominantly give high scores, consistently grading bad logically incorrect responses as ``very good.'' 



\section{Conclusion}
In this paper, we investigate how increasing levels of prompting impact the automatic evaluations made by LLMs-as-a-judge in measuring the quality of AI-generated text. We introduce a new taxonomy of quality criteria, summarizing commonly used criteria in automatic evaluations with LLMs into four broad categories: Content, Relevance, Integrity, and Engagement. We systematically evaluated several LLMs, including GPT-4, Llama-3, and others, across all settings of prompting to determine if more detailed instructions enhance the LLMs’ alignment with human judgements.  
Key findings include:  
\begin{itemize}
    \item Detailed quality criteria information might not be necessary in the most powerful models; for instance, GPT-4 shows a high level of agreement with human judgements even without detailed instruction.
    \item Simple perplexity values are very effective at estimating textual quality, often outperforming the results of prompting the LLMs-as-a-judge with basic instructions.
    \item Judging task-specific quality criteria like relevance or logical correctness requires more capable, larger models, aligning with previous research on the necessary model capabilities for critiquing~\cite{lin2024criticbenchbenchmarkingllmscritiquecorrect}.
\end{itemize}

\bigskip

\bibliography{evaluating_the_evaluator}

\end{document}